%% file: main_icml2020.tex
\documentclass{article}

\usepackage{microtype}
\usepackage{graphicx}
\usepackage{subfigure}
\usepackage{booktabs} 

\usepackage{amsmath,amssymb,amsfonts}

\usepackage{multirow}
\usepackage{siunitx}
\usepackage{graphicx}
\usepackage{tabularx}
\usepackage{float}
\usepackage{placeins}
\DeclareMathOperator*{\E}{\mathbb{E}}

\usepackage{hyperref}
\usepackage{cleveref}

\usepackage[accepted]{icml2020_sty/icml2020}

\begin{document}

\twocolumn[










\icmltitle{Bayesian Autoencoders: Analysing and Fixing the Bernoulli likelihood for Out-of-Distribution Detection}






\begin{icmlauthorlist}

\icmlauthor{Bang Xiang Yong}{ifm}
\icmlauthor{Tim Pearce}{ifm}
\icmlauthor{Alexandra Brintrup}{ifm}
\end{icmlauthorlist}

\icmlaffiliation{ifm}{Institute for Manufacturing, University of Cambridge, United Kingdom}

\icmlcorrespondingauthor{Bang Xiang Yong}{bxy20@cam.ac.uk}

\icmlkeywords{Machine Learning, ICML}

\vskip 0.3in
]



\printAffiliationsAndNotice{} 

\begin{abstract}

After an autoencoder (AE) has learnt to reconstruct one dataset, it might be expected that the likelihood on an out-of-distribution (OOD) input would be low. This has been studied as an approach to detect OOD inputs. Recent work showed this intuitive approach can fail for the dataset pairs FashionMNIST vs MNIST. This paper suggests this is due to the use of Bernoulli likelihood and analyses why this is the case, proposing two fixes: 1) Compute the \textit{uncertainty} of likelihood estimate by using a Bayesian version of the AE. 2) Use alternative distributions to model the likelihood.



\end{abstract}

\section{Introduction}
\label{submission}

Recent works by \cite{choi2018waic,daxberger2019bayesian,nalisnick2018do,ren2019likelihood} reported the unreliability of using the likelihood of generative models for out-of-distribution (OOD) detection. As such, many workarounds were developed such as typicality \cite{nalisnick2018do}, likelihood ratio \cite{ren2019likelihood} and Watanabe-Akaike Information Criterion (WAIC) \cite{choi2018waic}.

\begin{figure}[!htbp]
\vskip 0.2in
\begin{center}
\centerline{\includegraphics[width=\columnwidth]{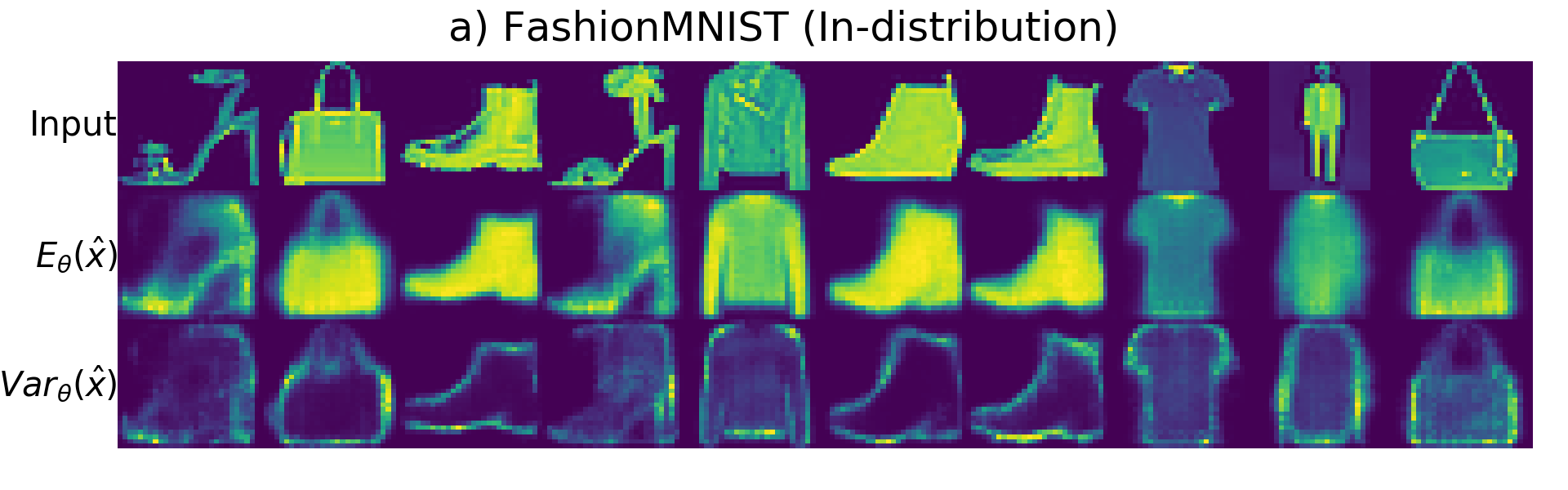}}
\centerline{\includegraphics[width=\columnwidth]{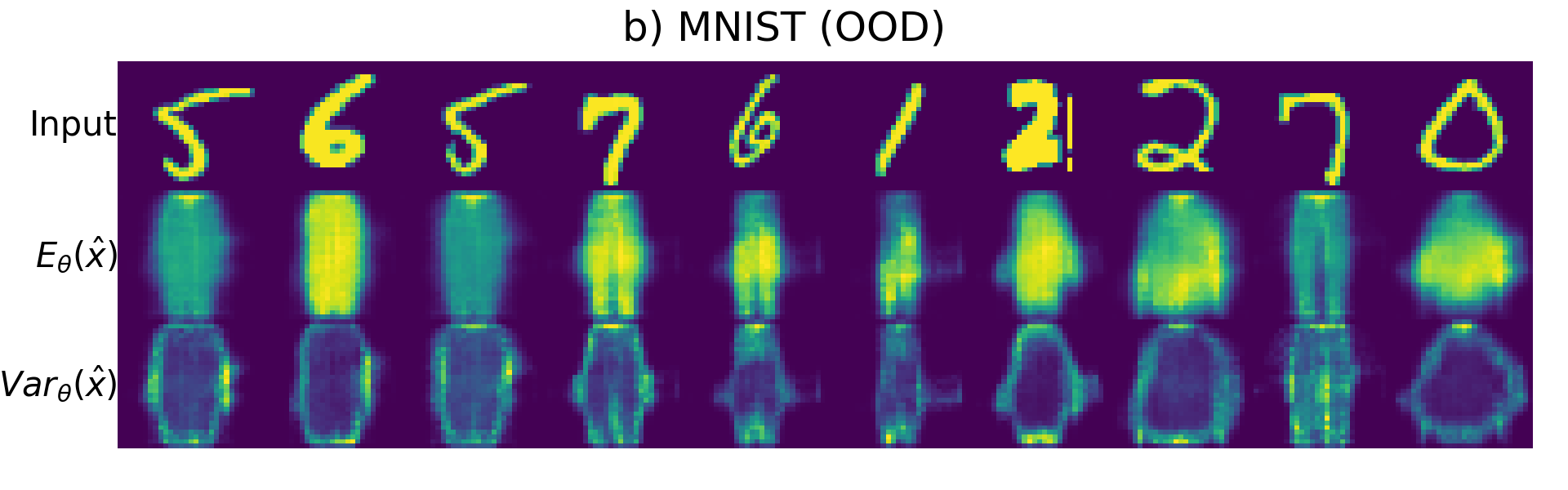}}
\centerline{\includegraphics[width=\columnwidth]{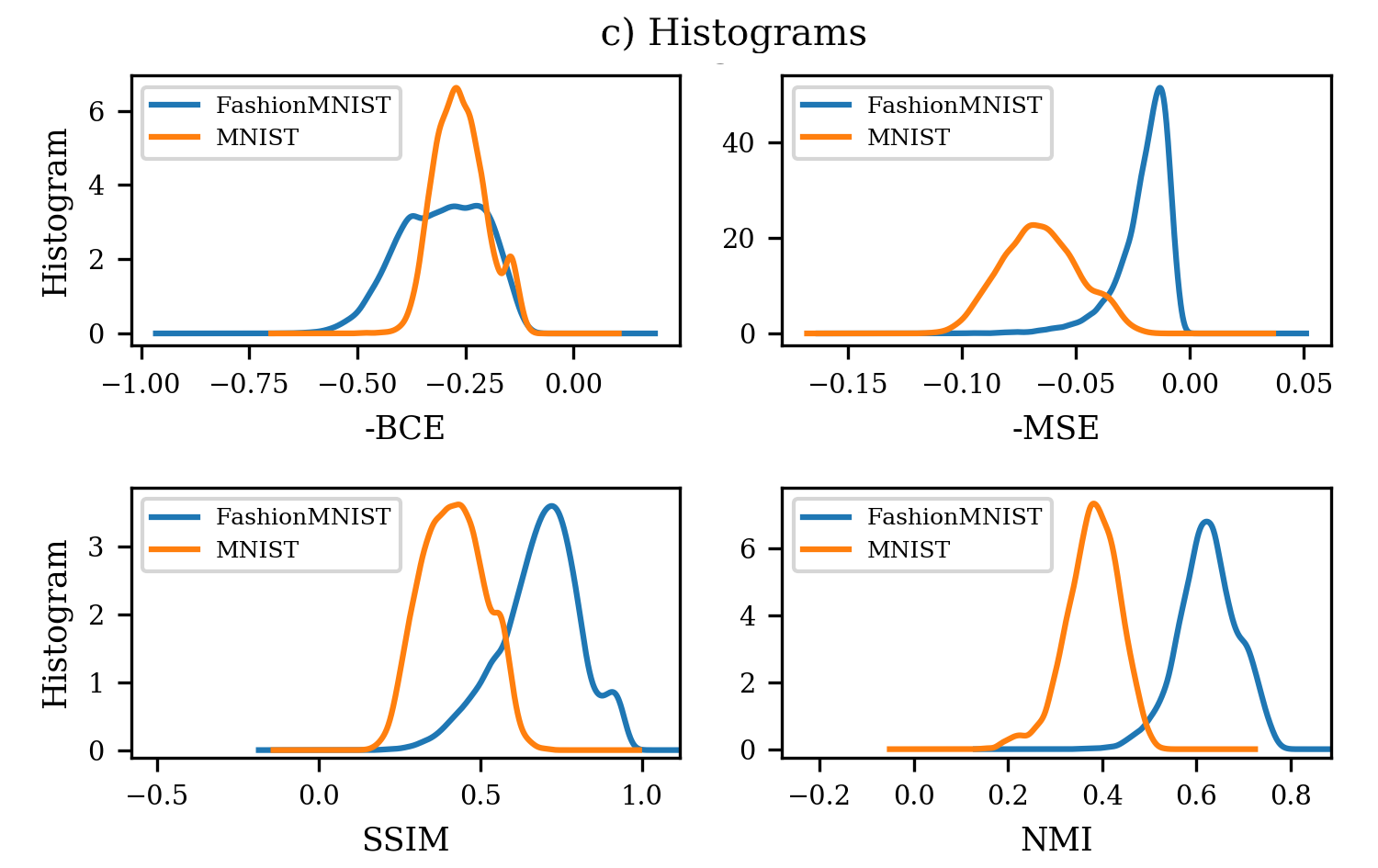}}
\caption{Mean of reconstructed signal $\E{}_{\theta}(\hat{\textbf{x}})$, and epistemic uncertainty of reconstructed signal $\mathrm{Var}_\theta(\hat{\textbf{x}}^*)$, on a) FashionMNIST (in-distribution) and b) MNIST (OOD). The reconstructed images of OOD data are qualitatively different from the input images as the model only knows about fashion images. Panel c) depicts histograms of image similarity measures: negative binary cross entropy (-BCE), negative mean-squared error (-MSE), structural similarity index (SSIM) and normalised mutual information (NMI). The higher the value, the more similar the reconstructed image is to the reference image. The samples are outputs from a VAE trained under Bernoulli likelihood.}
\label{fig-fashiomnist-samples}
\end{center}
\vskip -0.2in
\end{figure}

This unreliability is surprising, since autoencoders (AEs) have been extensively studied for OOD detection (also termed `anomaly detection') \cite{basora2019recent, dong2018threaded, kiran2018overview}. Intuitively, the idea is to train a neural network to reconstruct a set of training inputs, and during test time, we expect a higher `reconstruction loss` on inputs which deviate from the data distribution it was trained on. The likelihood is a probabilistic measure of the reconstruction loss, representing how well the input is recovered.




Figure~\ref{fig-fashiomnist-samples} shows a VAE trained on FashionMNIST and tested on both FashionMNIST and MNIST, and we observe the quality of reconstructed images of in-distribution and OOD data appears to be different; the model trained on FashionMNIST reconstructs what it knows (fashion images) when fed with digit images, and yet, the binary cross entropy (BCE)\footnote{The Bernoulli log-likelihood is equivalent to the binary cross entropy function.} is unable to tell the difference and assigns overlapping similarity score in Figure~\ref{fig-fashiomnist-samples}c). In contradiction, other well-studied image similarity measures \cite{wang2004image} assign lower scores on the OOD images. This implies the BCE is an unreliable measure for image similarity.

Motivated by this observation, we suggest that the culprit is in the likelihood. Specifically, we find recent papers which reported poor experimental results on OOD detection have used the Bernoulli likelihood in their variational autoencoder (VAE). Although using the Bernoulli likelihood is not coherent within a probabilistic framework for image data (support of likelihood is $\{$0,1$\}$ instead of [0,1] for each pixel value), this practice is common. For this reason, \cite{loaiza2019continuous} developed a fix, by including a normalisation constant and called this the Continuous Bernoulli distribution. Another widely used option is the Gaussian likelihood. Despite its support being not coherent with image data, empirically we find it to perform well for OOD.

In this paper, we develop Bayesian Autoencoders (BAE) and evaluate them on OOD detection using image dataset pairs which were shown to fail in recent studies. The contributions of this paper are:

\begin{enumerate}
  \item We show the poor OOD detection performance of Bernoulli likelihood is due to confounding by proportion of zeros in an image. Likewise, the Continuous Bernoulli likelihood suffers from similar issue. Furthermore, we demonstrate that BCE is an unreliable image similarity measure compared to other well-known image similarity measures. 
  \raggedbottom
  \item We motivate the development of the BAE for OOD detection, through various techniques to approximately sample from the posterior over weights. Although these methods have been explored with supervised BNNs, they are relatively understudied with AEs which are unsupervised models. 
  \raggedbottom
  \item We propose two simple ways fixes for the poor performance of Bernoulli likelihood for OOD detection on image datasets; 1) for stochastic AEs such as VAEs and BAEs, compute the uncertainty of log-likelihood estimate as the OOD score, and 2) use alternative distribution to model the likelihood.
\end{enumerate}

\section{Methods}

The AE is a neural network which maps a given set of unlabelled training data; $X = {\{\textbf{x}_1,\textbf{x}_2,\textbf{x}_3,...\textbf{x}_N\}}$, $\textbf{x}_{i} \in \rm I\!R^{D}$ into a set of reconstructed signals,  $\hat{X} = {\{\hat{\textbf{x}_1},\hat{\textbf{x}_2},\hat{\textbf{x}_3},...\hat{\textbf{x}_N}\}}$, $\hat{\textbf{x}_{i}} \in \rm I\!R^{D}$. An AE is parameterised by $\theta$, and consists of two parts: an encoder $f_\text{encoder}$, for mapping input data $\textbf{x}$ to a latent embedding, and a decoder $f_\text{decoder}$ for mapping the latent embedding to a reconstructed signal of the input $\hat{\textbf{x}}$ (i.e. $\hat{\textbf{x}} = {f_{\theta}}(\textbf{x}) = f_\text{decoder}(f_\text{encoder}(\textbf{x}))$) \cite{goodfellow2016deep}. 

Bayes' rule can be applied to the parameters of the AE, to create a BAE,
\begin{equation}\label{eq_posterior}
    p(\theta|X) = \frac{p(X|\theta)\  p(\theta)}{p(X)} \\ ,
\end{equation}
where $p(X|\theta)$ is the likelihood and $p(\theta)$ is the prior distribution of the AE parameters. During prediction with test data $\textbf{x}^*$, we are interested in the posterior predictive distribution, given by:
\begin{equation}\label{eq_predictive_full}
p(\hat{\textbf{x}}^*| \textbf{x}^*, X) = \int{p(\hat{\textbf{x}}^*| \textbf{x}^* , \theta)\ p(\theta|X)\ d\theta}
\end{equation} 

Since Equations~\ref{eq_posterior} and ~\ref{eq_predictive_full} are analytically intractable for a deep neural network, various approximate methods have been developed such as Stochastic Gradient Markov Chain Monte Carlo (SGHMC) \cite{chen2014stochastic}, MC-Dropout \cite{gal2016dropout}, Bayes by Backprop \cite{blundell2015weight}, and anchored ensembling \cite{pearce2018uncertainty} to sample from the posterior distribution. In contrast, a deterministic AE has its parameters estimated using maximum likelihood estimation (MLE) or maximum a posteriori (MAP) when regularisation is introduced. The VAE \cite{kingma2013auto} and BAE are AEs formulated differently within a probabilistic framework - in the VAE, only the latent embedding is stochastic while the $f_\text{encoder}$ and $f_\text{decoder}$ are deterministic while the BAE (similar to BNN) has distributions over all parameters of $f_\text{encoder}$ and $f_\text{decoder}$. The log-likelihood for a Bernoulli distribution is,
\begin{equation}\label{eq_bernoulli_lik}
\log{p(\textbf{x}|\theta)} = \frac{1}{D}\sum_{i=1}^{D}{x_i\cdot{\log{\hat{x_i}}}+(1-x_i)\cdot{\log{(1-\hat{x_i})}}},
\end{equation}
and for a diagonal Gaussian distribution,
\begin{equation} \label{eq_gaussian_loss}
\log{p(\textbf{x}|\theta)} = -(\frac{1}{D} \sum^{D}_{i=1}{\frac{1}{2\sigma_i^2}}(x_i-\hat{x_i})^2 + \frac{1}{2}\log{\sigma_i^2})
\end{equation} 
For a diagonal isotropic Gaussian likelihood with variance 1, minimising the mean-squared error (MSE) function is equivalent to minimising the negative log-likelihood (NLL).

Due to the intractability of Equation~\ref{eq_predictive_full}, we obtain a set of realised samples $\{\hat{\theta_t}\}^T_{t=1}$ from the posterior over weights and in a similar fashion to BNNs, we estimate the predictive mean $\E{}_{\theta}(\hat{\textbf{x}}^*)$ and the epistemic uncertainty of reconstructed signal $\mathrm{Var}_\theta(\hat{\textbf{x}}^*)$ as:
\begin{equation}\label{eq_predictive_mean}
\E{}_{\theta}(\hat{\textbf{x}}^*) = \frac{1}{T}\sum_{t=1}^{T}{f_{\hat{\theta_t}}(\textbf{x}^*)}
\end{equation}
\begin{equation}\label{eq_uncertainty_model_output}
\mathrm{Var}_\theta(\hat{\textbf{x}}^*) = \frac{1}{T}\sum_{t=1}^T{(f_{\hat{\theta_t}}(\textbf{x}^*) -\E{}_{\theta}(\hat{\textbf{x}}^*))^2}
\end{equation}
Furthermore, we can compute the mean and variance of the log-likelihood estimate, $\E{}_{\theta}(\log{p(\textbf{x}^*|\theta)})$ and $\text{Var}_\theta(\log{p(\textbf{x}^*|\theta)})$. The WAIC \cite{choi2018waic} is given by 
\begin{equation}\label{eq_waic}
\mathrm{WAIC}(\textbf{x}^*) = \E{}_{\theta}(\log{p(\textbf{x}^*|\theta)}) - \mathrm{Var}_\theta(\log{p(\textbf{x}^*|\theta)})
\end{equation}
\section{Experimental Setup}
\subsection{Models}
In our experiments, we train variants of AEs; deterministic AE, VAE and BAE\footnote{Code available:   \href{https://github.com/bangxiangyong/bae-ood-images}{github.com/bangxiangyong/bae-ood-images}}. For the BAE, we test several inference methods - MC-Dropout, Bayes by Backprop and anchored ensembling, always assuming a  diagonal isotropic Gaussian prior over weights. For all AEs, we trial three likelihoods - Bernoulli, Continuous Bernoulli or diagonal Gaussian with fixed variance 1. To tune the learning rate, we run a learning rate finder and employ a cyclic learning rate \cite{smith2017cyclical} during training of 20 epochs. For more details on the implementation, refer to Appendix~\ref{appendix-implementation}.

\subsection{Methods for OOD Detection}

We consider several methods to detect OOD inputs. For each AE type and likelihood, we compute $\mathrm{Var}_\theta(\hat{\textbf{x}}^*)$, expected value of log-likelihood $\E_{\theta}\mathrm{(LL)}$\footnote{For brevity, we denote LL = $\log{p(\textbf{x}^*|\theta)}$}, epistemic uncertainty of log-likelihood estimation, $\mathrm{Var_{\theta}(LL)}$ and WAIC on both the in-distribution and OOD test data. 

\section{Results and Discussion} \label{section-results}

In this section, firstly, following the evaluation by \cite{ren2019likelihood}, we analyse the confoundedness of each method due to the proportion of zeros in an image. Then, we evaluate and compare the methods of detecting OOD inputs using the area under the receiver operating curve (AUROC), a threshold agnostic measure which is common in recent literature \cite{choi2018waic, daxberger2019bayesian, nalisnick2018do, ren2019likelihood}.  

\begin{figure}[!ht]
\vskip 0.2in
\begin{center}
\centerline{\includegraphics[scale=0.8]{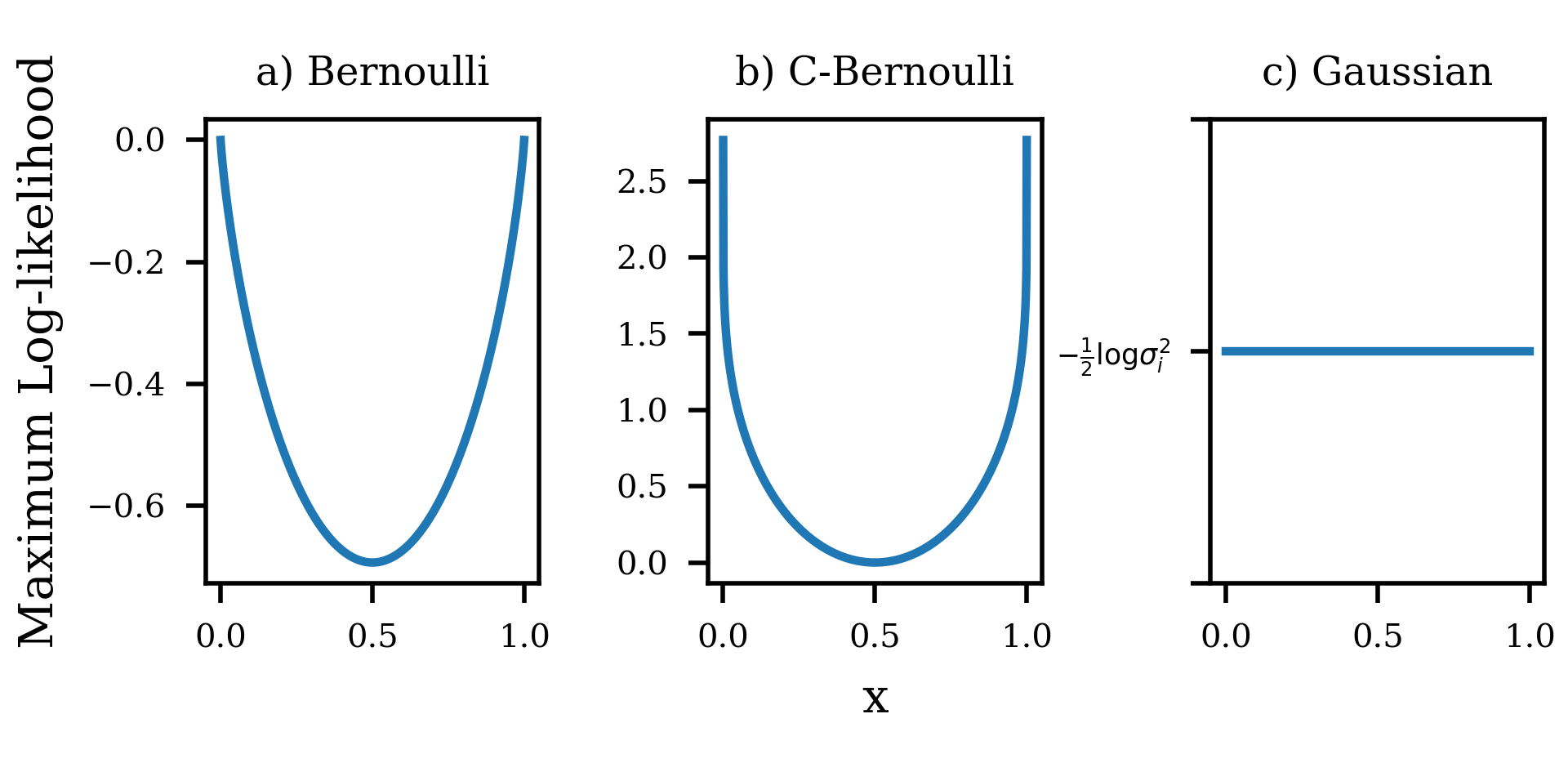}}
\caption{Maximum of a) Bernoulli and b) Continuous Bernoulli log-likelihood is influenced by the value of input, x. Conversely, c) maximum of Gaussian log-likelihood stays constant with respect to x.}
\label{fig-minmax-ll}
\end{center}
\vskip -0.2in
\end{figure}

\textbf{Confoundedness}. We now discuss an issue with using the Bernoulli likelihood as a method for OOD detection. Figure~\ref{fig-minmax-ll} plots the Bernoulli (Equation~\ref{eq_bernoulli_lik}), Continuous Bernoulli \cite{loaiza2019continuous} and Gaussian (Equation~\ref{eq_gaussian_loss}) log-likelihood for varying $x$, when $x=\hat{x}$ (i.e. the maximum value for a perfect reconstruction of the input). For the Bernoulli and Continuous Bernoulli likelihoods, we observe that the maximum value available is influenced by the value of $x$, with the highest log-likelihood available for when $x=0$ or $x=1$. 

This is a crucial observation. MNIST contains images with a high proportion of zeros, which means high log-likelihood scores are available. Conversely, FashionMNIST images have low proportion of zeros and therefore, low log-likelihood scores are available. This property is not exhibited by Gaussian log-likelihood, which maximum value remains constant when x is varied.

\begin{figure}[!ht]
\vskip 0.2in
\begin{center}
\centerline{\includegraphics[width=\columnwidth]{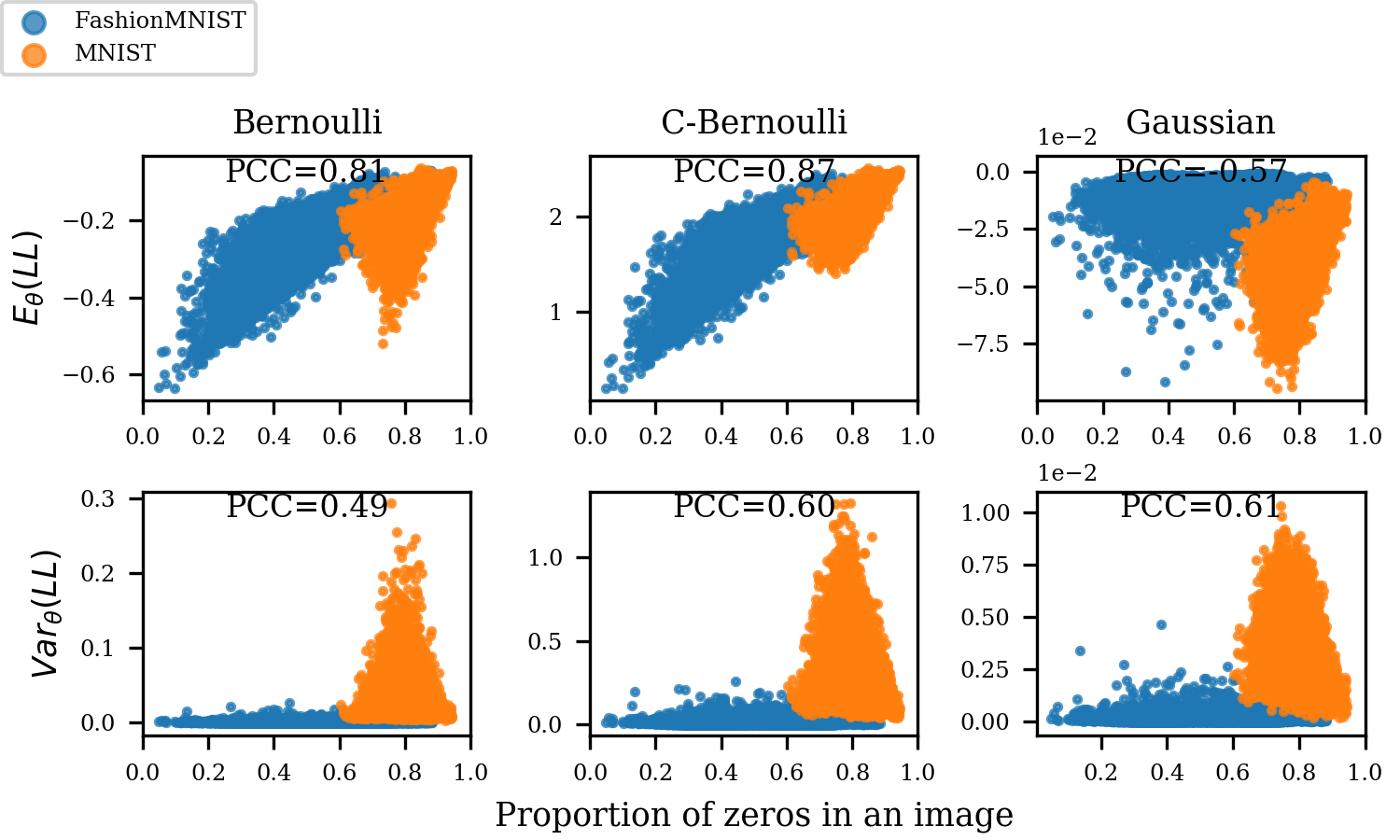}}
\caption{Relationship between $\E_{\theta}\mathrm{(LL)}$ (first row), $\mathrm{Var_{\theta}(LL)}$ (second row) and proportion of zeros in an image. Pearson correlation coefficient (PCC) for each plot is also shown. (FashionMNIST vs MNIST with BAE, Ensemble)}
\label{fig-pcc-ll}
\end{center}
\vskip -0.2in
\end{figure}

Furthermore, by plotting the correlation between proportion of zeros against $\E_{\theta}\mathrm{(LL)}$ and $\mathrm{Var_{\theta}(LL)}$ in Figure~\ref{fig-pcc-ll}, we find both the Bernoulli log-likelihood and its continuous variant are confounded by the proportion of zeros in the image (Pearson Correlation Coefficient 0.81, 0.87 respectively), in which the MNIST images have a larger proportion of zeros due to its background. This means the model assigns higher log-likelihoods to images with higher proportion of zeros, regardless of whether it is OOD input or not. In contrast, the $\E_{\theta}\mathrm{(LL)}$ for Gaussian likelihood and the $\mathrm{Var_{\theta}(LL)}$ do not suffer from such confoundedness. 


\input{tables/FashionMNIST_table}

\textbf{Poor OOD detection using Bernoulli likelihood.} Based on the results in Table~\ref{table-results-fashionMNIST}, by using the $\E_{\theta}\mathrm{(LL)}$ method, the performance is poorer for all models with Bernoulli likelihood (AUROC\textless{0.5}), compared to that of Gaussian likelihood (AUROC\textgreater{0.9})\footnote{Asymmetrically, however, the Bernoulli likelihood does perform well on MNIST vs FashionMNIST (Appendix~\ref{appendix-failure}, Table~\ref{table-results-MNIST}) which aligns with \cite{nalisnick2018do}}. We find the Continuous Bernoulli likelihood is unable to fix this poor performance. 

\textbf{Combining epistemic uncertainty and likelihood improves performance.} On its own, $\mathrm{Var}_\theta(\hat{\textbf{x}}^*)$ does not perform reliably for OOD detection, with the exception of anchored ensembling. This may be due to the quality of uncertainty which was reported to be poorer with MC-Dropout and variational inference, in comparison to ensembling \cite{yao2019quality, pearce2018uncertainty}. In spite of the report by \cite{nalisnick2018do} that ensembling is not robust towards OOD, we posit this would not happen if a Gaussian likelihood or the uncertainty estimate (either $\mathrm{Var}_\theta(\hat{\textbf{x}}^*)$ or $\mathrm{Var_{\theta}(LL)}$) was used. 

Surprisingly, when we combine the epistemic uncertainty and likelihood, which we call as the uncertainty of log-likelihood estimate, $\mathrm{Var_{\theta}(LL)}$, this method obtains good performance (AUROC\textgreater{0.9}) for all models (including VAE, BAE-MCDropout, and BAE-BayesBB which epistemic uncertainty performed poorly on their own). Moreover, for models which used Bernoulli likelihood, $\mathrm{Var_{\theta}(LL)}$ performed far better than $\E_{\theta}\mathrm{(LL)}$. This prevails even with the Continuous Bernoulli likelihood. 

Since $\mathrm{Var_{\theta}(LL)}$ performs well on its own, we question the need for WAIC since the AUROC using WAIC lies between that of $\E_{\theta}\mathrm{(LL)}$ and $\mathrm{Var_{\theta}(LL)}$ in most of the results.
Moreover, different from the implementation of \cite{choi2018waic} with the VAE, we obtain the epistemic uncertainty through importance sampling of a single model instead of an ensemble, which we find to be sufficient. 

For SVHN vs CIFAR10, we obtain good AUROC with one or more models (Appendix~\ref{appendix-add-results}, Table~\ref{table-results-SVHN}), however, we are unable to obtain the same for CIFAR10 vs SVHN and we offer a possible explanation in Appendix~\ref{appendix-failure}.

\section{Related Work}
\cite{choi2018waic, daxberger2019bayesian, nalisnick2018do, ren2019likelihood} reported the poor performance of using the log-likelihood method for OOD detection on FashionMNIST vs MNIST, which we find is due to choosing the Bernoulli distribution as likelihood, as the Gaussian likelihood do perform well on the same task. Although \cite{ren2019likelihood} first addressed the confoundedness of the likelihood, our work extends it by showing it is specific to the Bernoulli likelihood and conversely, the Gaussian likelihood and $\mathrm{Var_{\theta}(LL)}$ do not suffer from such issue. Our work is complementary to \cite{daxberger2019bayesian} who developed the Bayesian Variational Autoencoder (BVAE) with SGHMC, which we extend by exploring other approximation methods (MC-Dropout, Bayes by Backprop, anchored ensembling) common in BNN literature.


\section{Conclusion}

This paper investigated the reported inability of generative models to detect certain OOD inputs. We showed that this can be attributed to a problematic combination of the Bernoulli likelihood with images containing a high proportion of zeros.

The simplest fix for this is to switch to a Gaussian likelihood. Though the support of this distribution doesn't match the domain of pixel values, it nonetheless proved surprisingly effective (AUROC for all AEs improved from around 0.30 to 0.98). The second fix is to use the uncertainty of log-likelihood (for VAE and BAE). With the best inference techniques (anchored ensembling), AUROC was increased to 0.99. 


\section*{Acknowledgements}

The research presented was supported by EMPIR (European Metrology Programme for Innovation and Research) under the MET4FOF (Metrology for the Factory of the Future) project, as well as the PITCH-IN (Promoting the Internet of Things via Collaborations between HEIs and Industry) project funded by Research England. 

\bibliography{icml2020_sty/main_icml2020}
\bibliographystyle{icml2020_sty/icml2020}

\include{appendix}

\end{document}

%% file: tables/FashionMNIST_table.tex
\begin{table}[!ht]
\centering
\caption{(FashionMNIST vs MNIST) AUROC for detecting OOD inputs with deterministic AE, VAE and BAE (MC-Dropout, Bayes by Backprop and anchored ensembling) with posterior approximated under various techniques. For each likelihood- Bernoulli, Continuous Bernoulli and Gaussian, we compare the performance of using $\mathrm{Var}_\theta(\hat{\textbf{x}})$, $\E_{\theta}\mathrm{(LL)}$, $\mathrm{Var_{\theta}(LL)}$, and WAIC. Results where AUROC\textgreater{0.8} are in bold.}
\label{table-results-fashionMNIST}

\resizebox{.48\textwidth}{!}{
\begin{tabular}{@{}llcccc@{}}
\toprule
\multicolumn{1}{c}{\multirow{2}{*}{Model}} & \multicolumn{1}{c}{\multirow{2}{*}{Likelihood}} & \multicolumn{4}{c}{AUROC}                                  \\ \cmidrule(l){3-6} 
\multicolumn{1}{c}{}                       & \multicolumn{1}{c}{}                            & $\mathrm{Var}_\theta(\hat{\textbf{x}}^*)$ & $\E_{\theta}\mathrm{(LL)}$ & $\mathrm{Var_{\theta}(LL)}$ & WAIC \\ \midrule
Deterministic AE                                         & Ber($\hat{{x}}^*$)                                             & -                & 0.230       & -           & -           \\
Deterministic AE                                         & C-Ber($\hat{{x}}^*$)                                           & -                & 0.203       & -           & -           \\
Deterministic AE                                         & N($\hat{{x}}^*$,1)                                               & -                & \textbf{0.978}       & -           & -           \\ \midrule
VAE                                        & Ber($\hat{{x}}^*$)                                             & 0.629            & 0.393       & \textbf{0.982}       & 0.454       \\
VAE                                        & C-Ber($\hat{{x}}^*$)                                           & 0.365            & 0.301       & \textbf{0.995}       & 0.389       \\
VAE                                        & N($\hat{{x}}^*$,1)                                               & 0.420            & \textbf{0.987}       & \textbf{0.948}       & \textbf{0.987}      \\ \midrule
BAE, MC-Dropout                            & Ber($\hat{{x}}^*$)                                             & 0.610            & 0.224       & \textbf{0.983}       & 0.250       \\
BAE, MC-Dropout                            & C-Ber($\hat{{x}}^*$)                                           & 0.670            & 0.183       & \textbf{0.981}       & 0.237       \\
BAE, MC-Dropout                            & N($\hat{{x}}^*$,1)                                               & \textbf{0.820}            & \textbf{0.977}  & \textbf{0.991}       & \textbf{0.978}       \\ \midrule
BAE, BayesBB                                    & Ber($\hat{{x}}^*$)                                             & 0.736            & 0.258       & \textbf{0.987}       & 0.371       \\
BAE, BayesBB                                    & C-Ber($\hat{{x}}^*$)                                           & 0.750            & 0.204       & \textbf{0.991}       & 0.308       \\
BAE, BayesBB                                    & N($\hat{{x}}^*$,1)                                               & 0.701            & \textbf{0.945}       & \textbf{0.941}       & \textbf{0.951}       \\ \midrule
BAE, Ensemble                              & Ber($\hat{{x}}^*$)                                             & \textbf{0.990}            & 0.230       & \textbf{0.999}       & 0.314       \\
BAE, Ensemble                              & C-Ber($\hat{{x}}^*$)                                           & \textbf{0.978}            & 0.191       & \textbf{0.999}       & 0.418       \\
BAE, Ensemble                              & N($\hat{{x}}^*$,1)                                               & \textbf{0.995}            & \textbf{0.976}       & \textbf{0.998}       & \textbf{0.980}       \\ \bottomrule
\end{tabular}
}
\end{table}

%% file: appendix.tex
\appendix

\input{appendix_implementations}

\section{Additional results} \label{appendix-add-results}
Results for other dataset pairs (MNIST vs FashionMNIST, SVHN vs CIFAR10, and CIFAR10 vs SVHN) are shown here. 

\raggedbottom
\input{tables/MNIST_table}
\raggedbottom
\input{tables/SVHN_table}
\raggedbottom
\input{tables/CIFAR10_table}
\raggedbottom

\section{Failure mode on CIFAR10 vs SVHN} \label{appendix-failure}

\begin{figure}[H]
\vskip 0.2in
\begin{center}
\centerline{\includegraphics[width=\columnwidth]{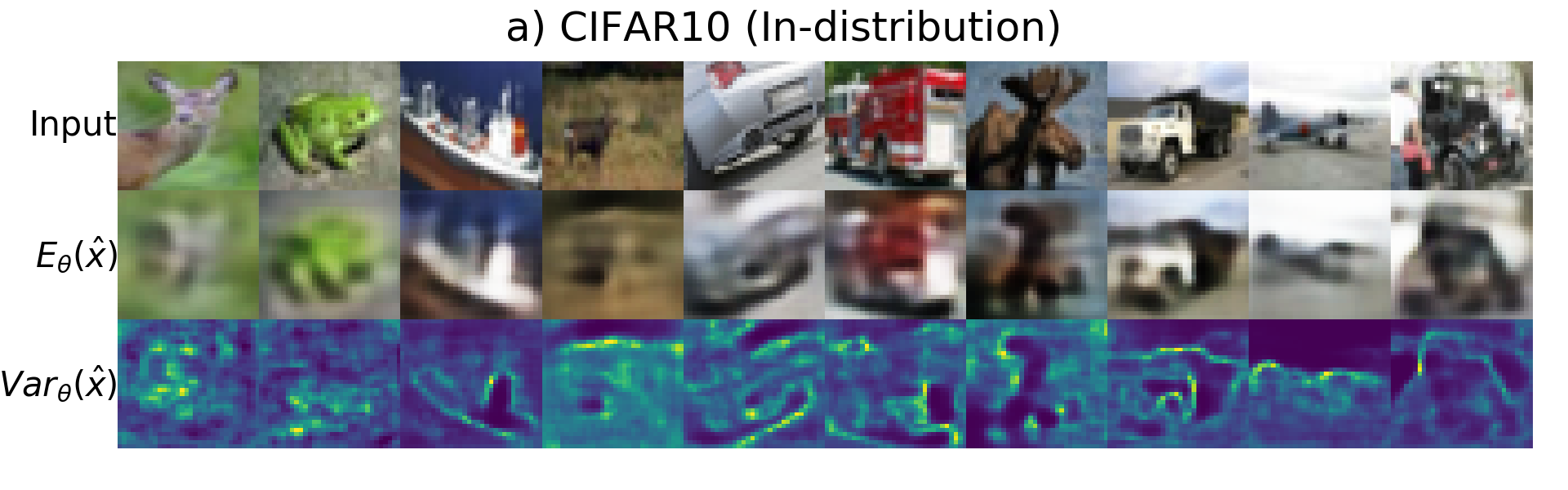}}
\centerline{\includegraphics[width=\columnwidth]{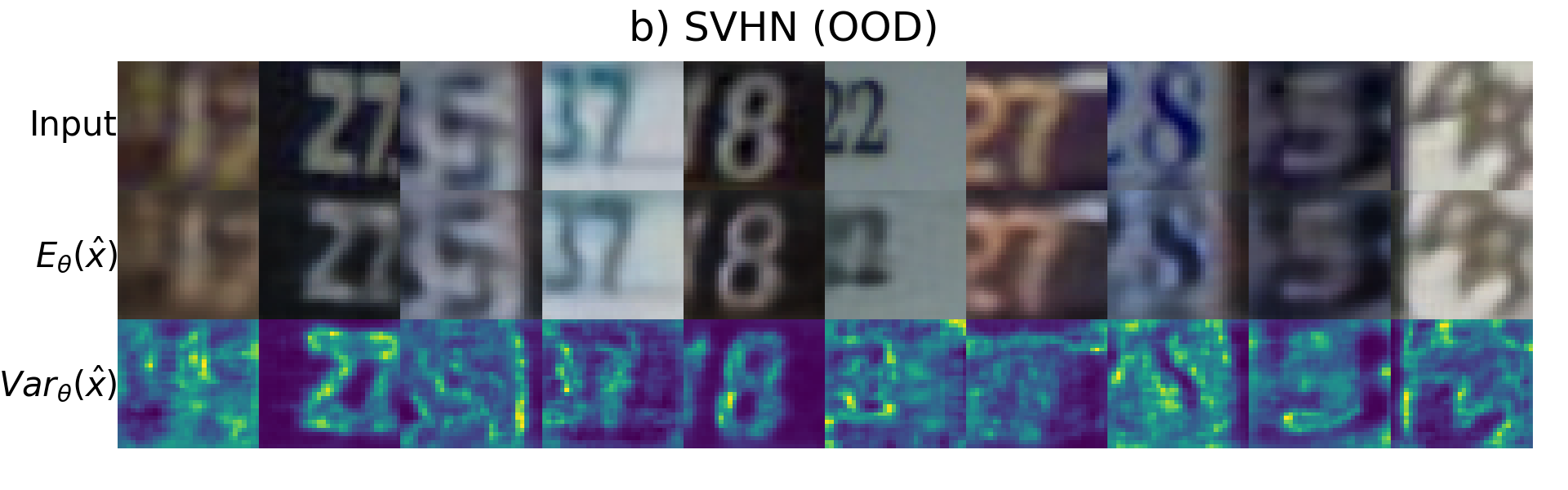}}
\caption{(CIFAR10 vs SVHN) Mean and uncertainty of reconstructed signals. In this failure mode, the reconstructed images of OOD data closely resemble the input images, which indicate the model has learnt to merely copy the inputs instead of learning the semantic representation of the training distribution.}
\label{fig-cifar-samples}
\end{center}
\vskip -0.2in
\end{figure}

We were unable to obtain any good results for this dataset pair (Table~\ref{table-results-CIFAR10}). In this scenario, we posit that the models are unable to learn meaningful representation of the training distribution. In turn, they have learnt to `copy`, as we examine the reconstructed images which are similar to the inputs, even on OOD dataset (Figure~\ref{fig-cifar-samples}). When this occurs, even the $\mathrm{Var_\theta(\hat{\textbf{x}}^*)}$ and Var(LL) are unable to detect OOD inputs. Thus, as a cautionary note: although the AEs can reconstruct the in-distribution data with minimal error, they may not be immediately reliable for OOD detection. Even with sparsity and regularisation in the AEs, we were unable to circumvent this.

\onecolumn
\section{Reconstructed images of test sets}


\begin{figure}[!htbp]
\centering     
\subfigure{\label{fig:a}\includegraphics[width=80mm]{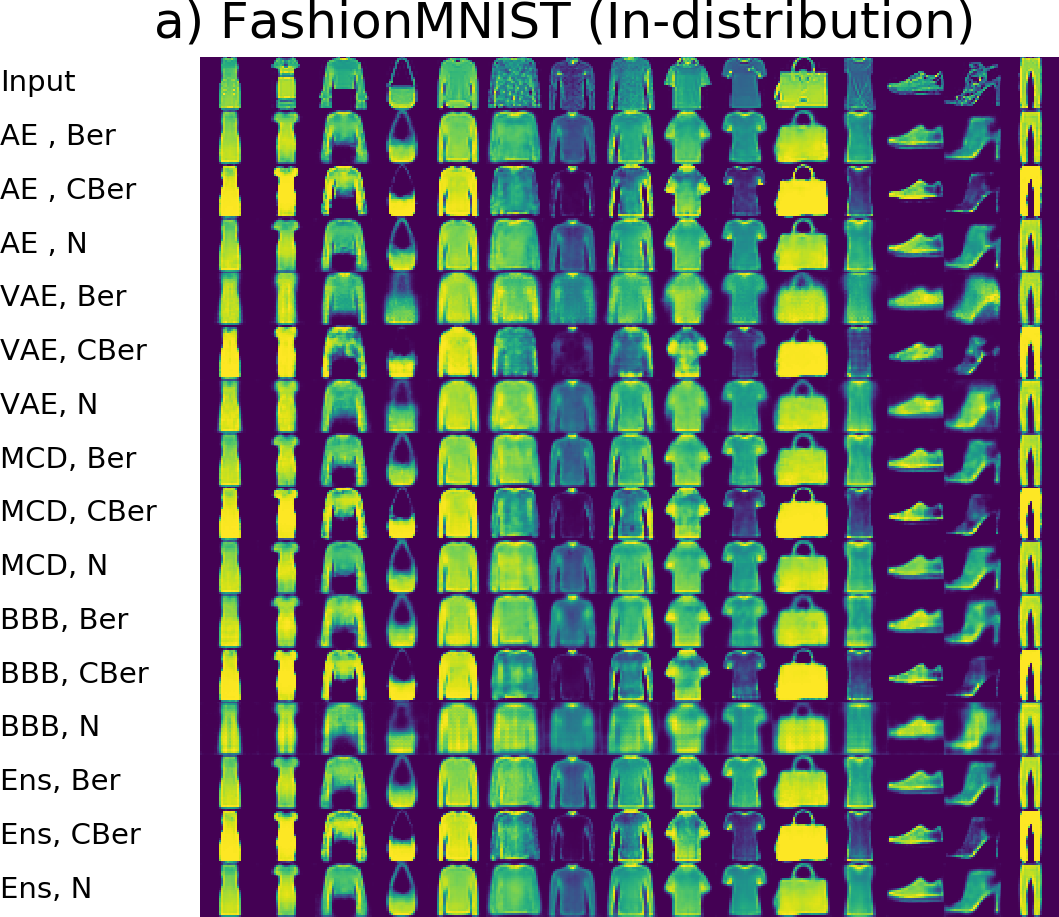}}
\subfigure{\label{fig:b}\includegraphics[width=80mm]{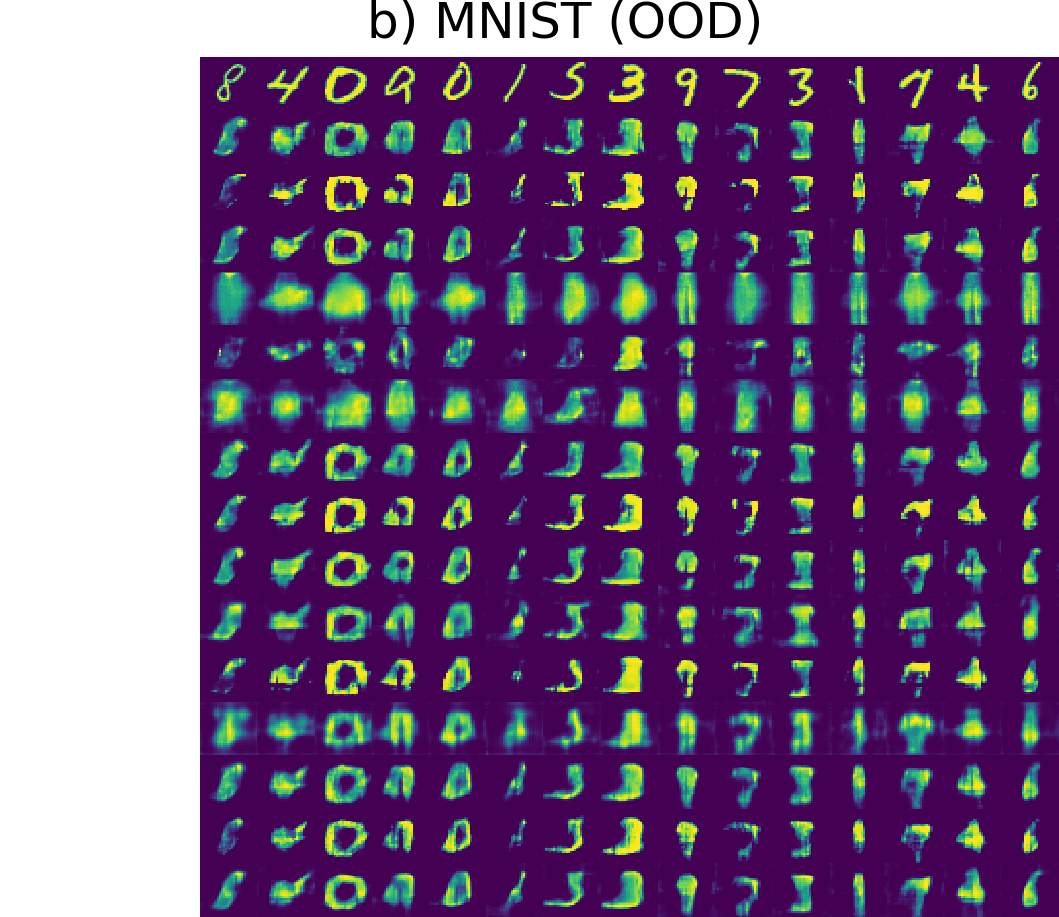}}
\caption{(FashionMNIST vs MNIST) Mean of reconstructed images for each model and likelihood.}
\end{figure}

\begin{figure}[!htbp]
\centering     
\subfigure{\label{fig:a}\includegraphics[width=80mm]{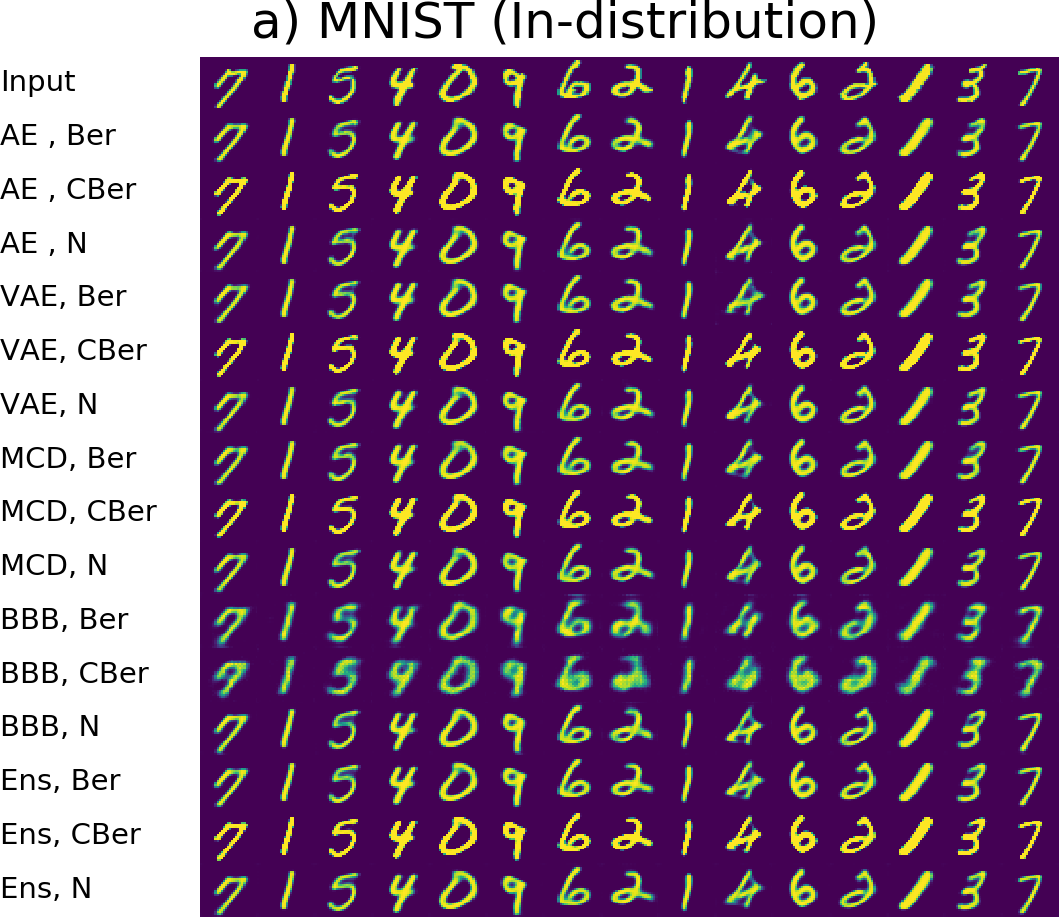}}
\subfigure{\label{fig:b}\includegraphics[width=80mm]{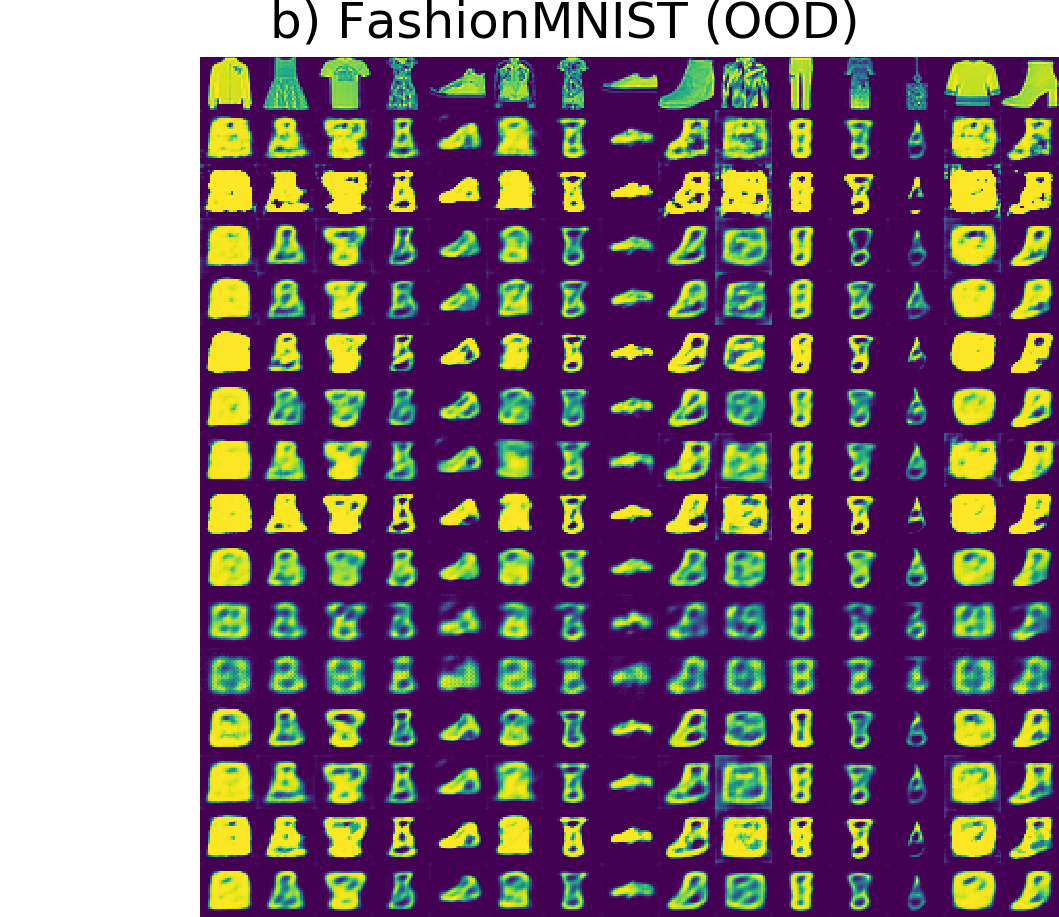}}
\caption{(MNIST vs FashionMNIST) Mean of reconstructed images for each model and likelihood.}
\end{figure}

\begin{figure}[!htbp]
\centering     
\subfigure{\label{fig:a}\includegraphics[width=80mm]{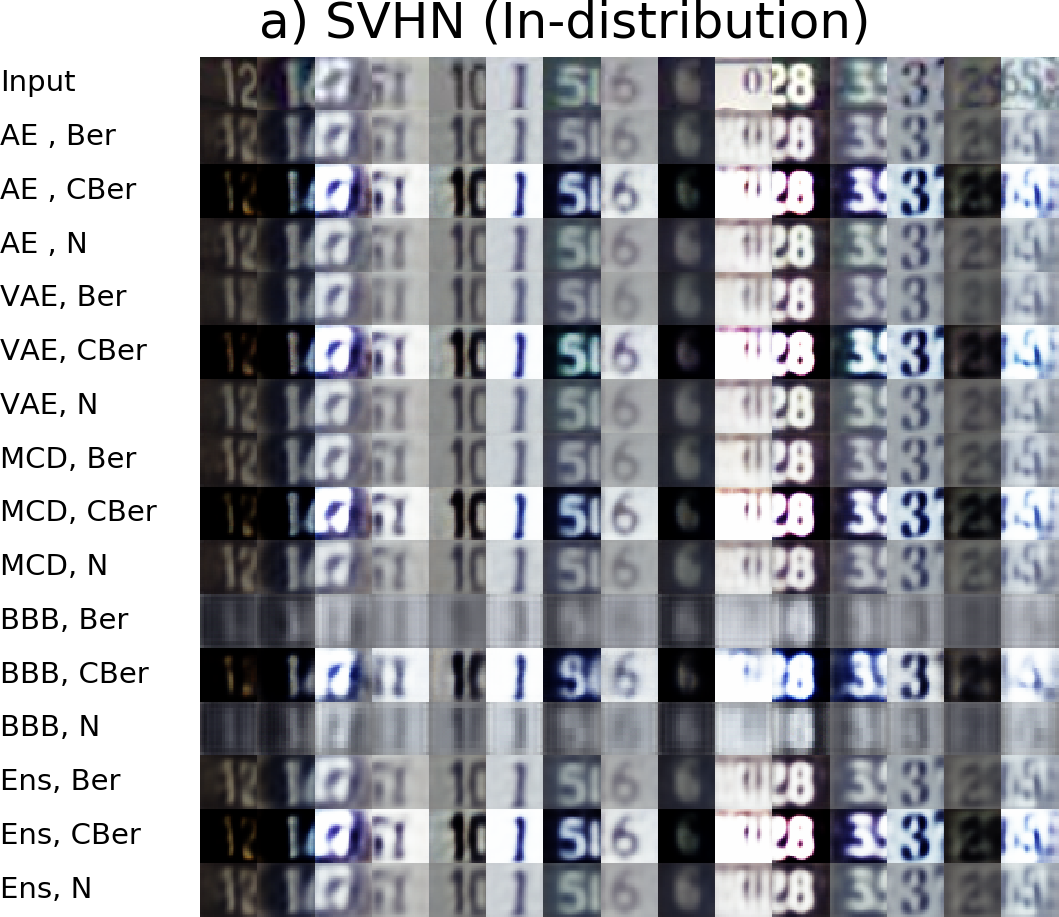}}
\subfigure{\label{fig:b}\includegraphics[width=80mm]{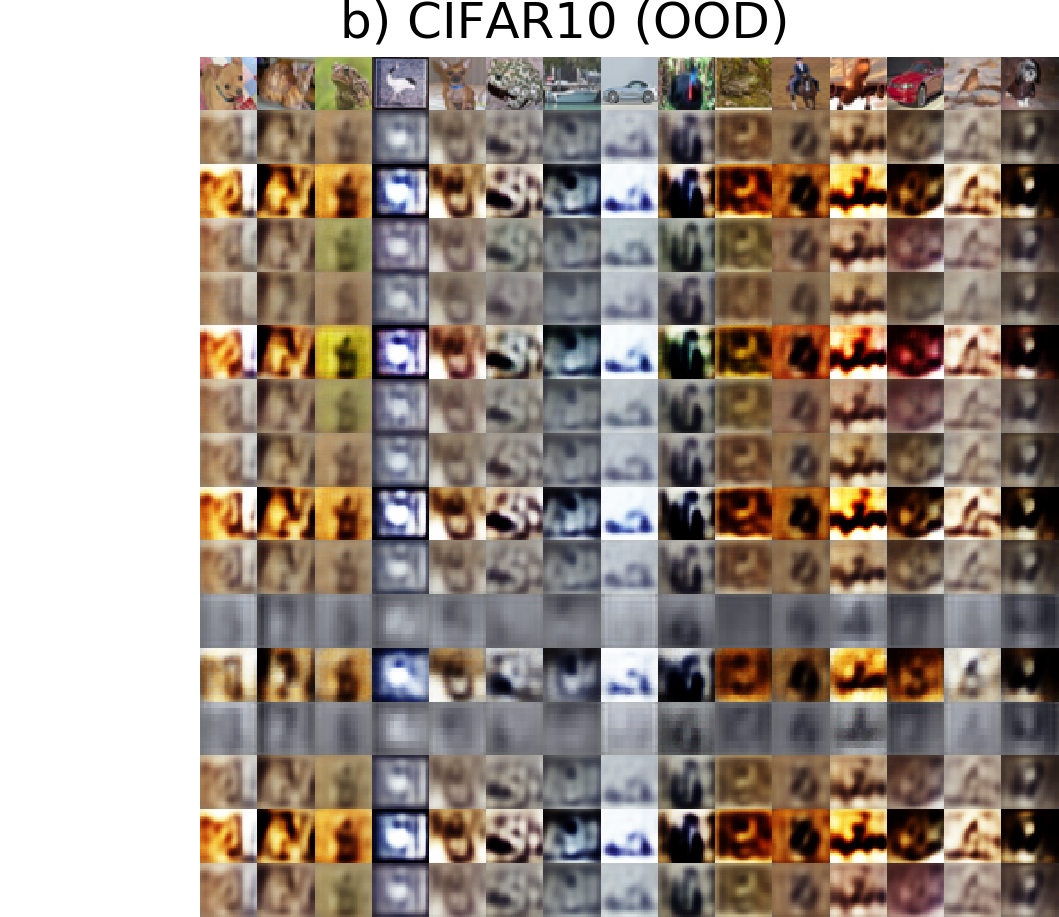}}
\caption{(SVHN vs CIFAR10) Mean of reconstructed images for each model and likelihood.}
\end{figure}

\begin{figure}[!htbp]
\centering     
\subfigure{\label{fig:a}\includegraphics[width=80mm]{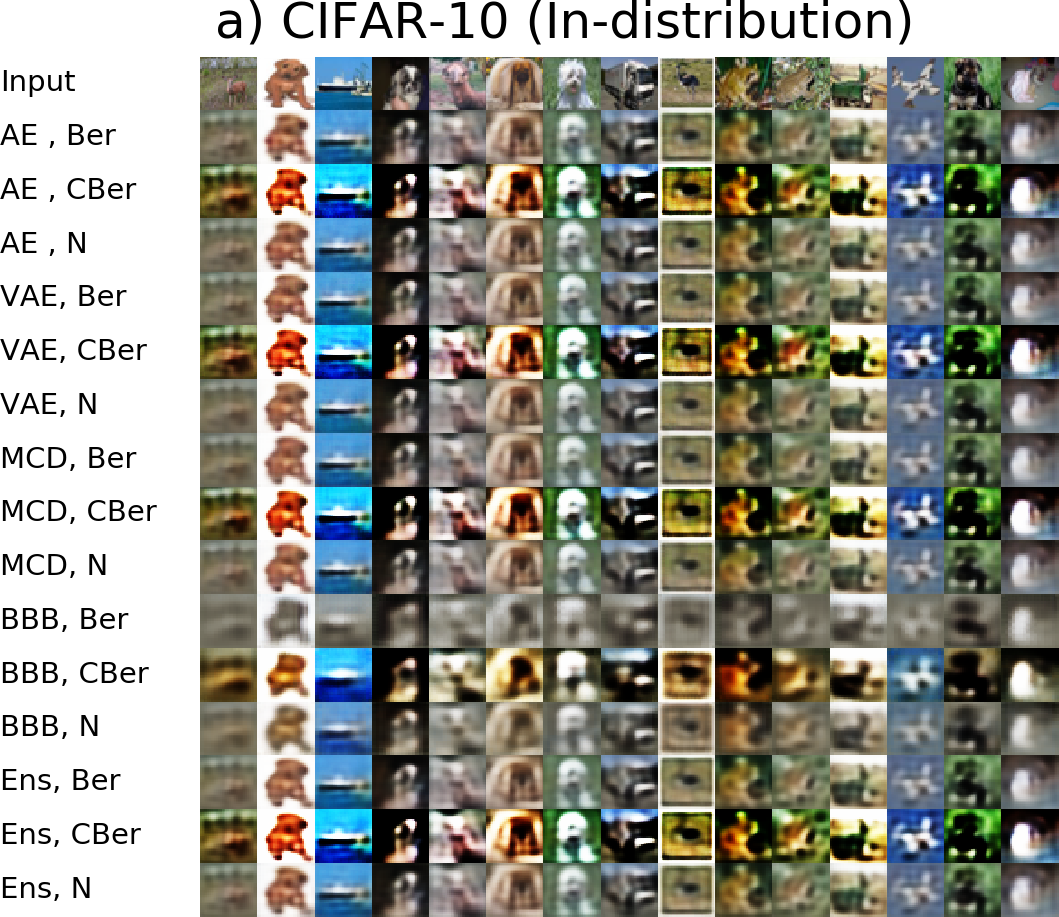}}
\subfigure{\label{fig:b}\includegraphics[width=80mm]{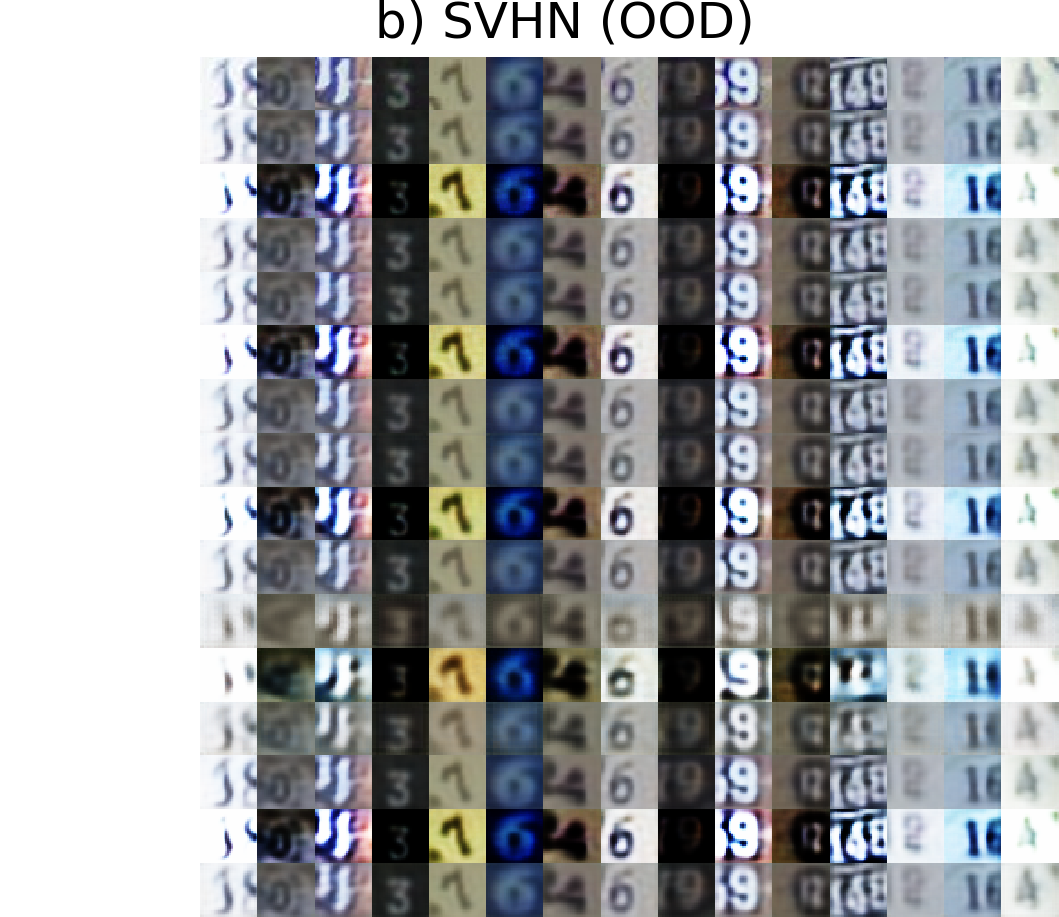}}
\caption{(CIFAR10 vs SVHN) Mean of reconstructed images for each model and likelihood.}
\end{figure}



%% file: appendix_implementations.tex
\section{Implementation details} \label{appendix-implementation}

Our code is available at \href{https://github.com/bangxiangyong/bae-ood-images}{github.com/bangxiangyong/bae-ood-images} for reproducibility. We use the default dataset splits provided by \textit{torchvision}\footnote{https://pytorch.org/docs/stable/torchvision}. We test the trained model on the test split for both in-distribution and OOD inputs. 

\subsection{Network architectures}

For each AE, we control the network architecture for a fair comparison. We use leaky ReLu as the activation function for each network in our experiments with a slope of 0.01. We apply the sigmoid function on the outputs. The latent size for FashionMNIST, MNIST, SVHN and CIFAR10 (in-distribution) datasets are 20, 20, 50 and 100 respectively.

\input{tables/architectures_table}

\subsection{Learning rate}

We run a learning rate finder and employ a sawtooth cyclic learning rate protocol \cite{smith2017cyclical} with Adam optimiser. With a batch size of 100, the training lasts for 20 epochs. We swept the value for scaling the regularisation term over values of $\{10, 2, 1, 0.1, 0.01, 0.001\}$ for each combination of model and likelihood.

\subsection{Number of samples}

We train 5 independently initialised AEs for BAE-Ensemble. With the VAE, BAE-MCDropout and BAE-BayesBB, we draw 100 samples from the posterior for each input image.


%% file: tables/architectures_table.tex
\begin{table}[H]
\caption{Encoder architectures for FashionMNIST, MNIST, CIFAR10 and SVHN acting as in-distribution training dataset. The decoder is a reflection of the encoder, in which the convolution layers are replaced by convolution transpose layers.}
\label{table-architecture}
\centering
\vskip 0.1in
a) FashionMNIST and MNIST
\begin{tabular}{@{}cccc@{}}
\toprule
Layer       & Output dimensions & Kernel & Strides \\ \midrule
Convolution & 13 x 13 x 32      & 4 x 4  & 2 x 2   \\
Convolution & 10 x 10 x 64      & 4 x 4  & 1 x 1   \\
Reshape     & 6400              & -      & -       \\
Dense       & Latent size       & -      & -      \\ \bottomrule
\end{tabular}
\newline \newline
b) CIFAR10 and SVHN
\begin{tabular}{@{}cccc@{}}
\toprule
Layer       & Output dimensions & Kernel & Strides \\ \midrule
Convolution & 15 x 15 x 32      & 4 x 4  & 2 x 2   \\
Convolution & 12 x 12 x 64      & 4 x 4  & 1 x 1   \\
Convolution & 5 x 5 x 128       & 4 x 4  & 2 x 2   \\
Reshape     & 3200              & -      & -       \\
Dense       & Latent size       & -      & -      \\ \bottomrule
\end{tabular}
\end{table}

%% file: tables/MNIST_table.tex
\begin{table}[ht]
\caption{(MNIST vs FashionMNIST) AUROC for various models and likelihoods. Results where AUROC\textgreater{0.8} are in bold.}
\label{table-results-MNIST}
\resizebox{.48\textwidth}{!}{
\begin{tabular}{@{}llcccc@{}}
\toprule
\multicolumn{1}{c}{\multirow{2}{*}{Model}} & \multicolumn{1}{c}{\multirow{2}{*}{Likelihood}} & \multicolumn{4}{c}{AUROC}                                  \\ \cmidrule(l){3-6} 
\multicolumn{1}{c}{}                       & \multicolumn{1}{c}{}                            & $\mathrm{Var}_\theta(\hat{\textbf{x}}^*)$ & $\E_{\theta}\mathrm{(LL)}$ & $\mathrm{Var_{\theta}(LL)}$ & WAIC \\ \midrule
Deterministic AE                                         & Ber($\hat{x}^*$)                                             & -                & \textbf{1.00}        & -           & -           \\
Deterministic AE                                         & C-Ber($\hat{x}^*$)                                           & -                & \textbf{1.00}        & -           & -           \\
Deterministic AE                                         & N($\hat{x}^*$,1)                                               & -                & \textbf{0.999}       & -           & -           \\ \midrule
VAE                                        & Ber($\hat{x}^*$)                                             & \textbf{0.996}            & \textbf{1.00}        & \textbf{1.00}        & \textbf{1.00}        \\
VAE                                        & C-Ber($\hat{x}^*$)                                           & \textbf{0.990}            & \textbf{1.00}        & \textbf{0.997}       & \textbf{1.00}        \\
VAE                                        & N($\hat{x}^*$,1)                                               & \textbf{0.973}            & \textbf{0.999}       & \textbf{0.982}       & \textbf{0.999}       \\ \midrule
BAE, MC-Dropout                            & Ber($\hat{x}^*$)                                             & 0.600            & \textbf{1.00}        & \textbf{0.995}       & \textbf{1.00}        \\
BAE, MC-Dropout                            & C-Ber($\hat{x}^*$)                                           & 0.713            & \textbf{1.00}        & \textbf{0.930}       & \textbf{0.999}       \\
BAE, MC-Dropout                            & N($\hat{x}^*$,1)                                               & 0.331            & \textbf{0.999}       & 0.636       & \textbf{0.999}       \\ \midrule
BAE, BayesBB                                    & Ber($\hat{x}^*$)                                             & \textbf{0.834}            & \textbf{0.998}       & \textbf{0.885}       & \textbf{0.921}       \\
BAE, BayesBB                                    & C-Ber($\hat{x}^*$)                                           & \textbf{0.864}            & \textbf{0.990}       & 0.714       & \textbf{0.904}       \\
BAE, BayesBB                                    & N($\hat{x}^*$,1)                                               & \textbf{0.859}            & \textbf{0.998}       & \textbf{0.863}       & \textbf{0.998}       \\ \midrule
BAE, Ensemble                              & Ber($\hat{x}^*$)                                             & \textbf{0.995}            & \textbf{1.00}        & \textbf{0.999}       & \textbf{1.00}        \\
BAE, Ensemble                              & C-Ber($\hat{x}^*$)                                           & \textbf{0.996}            & \textbf{1.00}        & \textbf{0.997}       & \textbf{1.00}        \\
BAE, Ensemble                              & N($\hat{x}^*$,1)                                               & \textbf{0.996}            & \textbf{0.999}       & \textbf{0.991}       & \textbf{0.999}       \\ \bottomrule
\end{tabular}
}
\end{table}

%% file: tables/SVHN_table.tex
\begin{table}[ht]
\caption{(SVHN vs CIFAR10) AUROC for various models and likelihoods. Results where AUROC\textgreater{0.8} are in bold.}
\label{table-results-SVHN}
\resizebox{.48\textwidth}{!}{
\begin{tabular}{@{}llcccc@{}}
\toprule
\multicolumn{1}{c}{\multirow{2}{*}{Model}} & \multicolumn{1}{c}{\multirow{2}{*}{Likelihood}} & \multicolumn{4}{c}{AUROC}                                  \\ \cmidrule(l){3-6} 
\multicolumn{1}{c}{}                       & \multicolumn{1}{c}{}                            & $\mathrm{Var_\theta(\hat{\textbf{x}}^*)}$ & $\E_{\theta}\mathrm{(LL)}$ & $\mathrm{Var_{\theta}(LL)}$ & WAIC \\ \midrule
Deterministic AE                                         & Ber($\hat{x}^*$)                                              & -                & 0.442       & -           & -           \\
Deterministic AE                                         & C-Ber($\hat{x}^*$)                                           & -                & 0.436       & -           & -           \\
Deterministic AE                                         & N($\hat{x}^*$,1)                                               & -                & \textbf{0.972}       & -           & -           \\ \midrule
VAE                                        & Ber($\hat{x}^*$)                                             & 0.751            & 0.455       & \textbf{0.948}       & 0.456       \\
VAE                                        & C-Ber($\hat{x}^*$)                                           & 0.526            & 0.435       & \textbf{0.947}       & 0.436       \\
VAE                                        & N($\hat{x}^*$,1)                                               & 0.630            & \textbf{0.973}       & \textbf{0.964}       & \textbf{0.973}       \\ \midrule
BAE, MC-Dropout                            & Ber($\hat{x}^*$)                                             & \textbf{0.820}            & 0.445       & \textbf{0.923}       & 0.446       \\
BAE, MC-Dropout                            & C-Ber($\hat{x}^*$)                                           & \textbf{0.861}            & 0.441       & \textbf{0.898}       & 0.445       \\
BAE, MC-Dropout                            & N($\hat{x}^*$,1)                                               & \textbf{0.802}            & \textbf{0.967}       & \textbf{0.938}       & \textbf{0.967}       \\ \midrule
BAE, BayesBB                                    & Ber($\hat{x}^*$)                                             & 0.649            & 0.570       & 0.797       & 0.606       \\
BAE, BayesBB                                    & C-Ber($\hat{x}^*$)                                           & \textbf{0.855}            & 0.485       & \textbf{0.877}       & 0.502       \\
BAE, BayesBB                                    & N($\hat{x}^*$,1)                                               & 0.760            & \textbf{0.918}       & \textbf{0.851}       & \textbf{0.915}       \\ \midrule
BAE, Ensemble                              & Ber($\hat{x}^*$)                                             & \textbf{0.922}            & 0.440       & \textbf{0.950}       & 0.441       \\
BAE, Ensemble                              & C-Ber($\hat{x}^*$)                                           & \textbf{0.921}            & 0.436       & \textbf{0.939}       & 0.439       \\
BAE, Ensemble                              & N($\hat{x}^*$,1)                                               & \textbf{0.931}            & \textbf{0.969}       & \textbf{0.960}       & \textbf{0.969}       \\ \bottomrule
\end{tabular}
}
\end{table}

%% file: tables/CIFAR10_table.tex
\begin{table}[H]
\caption{(CIFAR10 vs SVHN) AUROC for various models and likelihoods. Results where AUROC\textgreater{0.6} are in bold.}
\label{table-results-CIFAR10}
\resizebox{.48\textwidth}{!}{
\begin{tabular}{@{}llcccc@{}}
\toprule
\multicolumn{1}{c}{\multirow{2}{*}{Model}} & \multicolumn{1}{c}{\multirow{2}{*}{Likelihood}} & \multicolumn{4}{c}{AUROC}                                  \\ \cmidrule(l){3-6} 
\multicolumn{1}{c}{}                       & \multicolumn{1}{c}{}                            & $\mathrm{Var}_\theta(\hat{\textbf{x}}^*)$ & $\E_{\theta}\mathrm{(LL)}$ & $\mathrm{Var_{\theta}(LL)}$ & WAIC \\ \midrule
Deterministic AE                                         & Ber($\hat{x}^*$)                                             & -                & \textbf{0.601}       & -           & -           \\
Deterministic AE                                         & C-Ber($\hat{x}^*$)                                          & -                & \textbf{0.607}       & -           & -           \\
Deterministic AE                                         & N($\hat{x}^*$,1)                                               & -                & 0.038       & -           & -           \\ \midrule
VAE                                        & Ber($\hat{x}^*$)                                             & 0.492            & \textbf{0.604}       & 0.051       & \textbf{0.604}       \\
VAE                                        & C-Ber($\hat{x}^*$)                                           & 0.518            & \textbf{0.609}       & 0.077       & \textbf{0.609}       \\
VAE                                        & N($\hat{x}^*$,1)                                               & 0.522            & 0.045       & 0.048       & 0.045       \\ \midrule
BAE, MC-Dropout                            & Ber($\hat{x}^*$)                                             & 0.241            & \textbf{0.600}       & 0.113       & 0.599       \\
BAE, MC-Dropout                            & C-Ber($\hat{x}^*$)                                           & 0.214            & \textbf{0.607}       & 0.144       & \textbf{0.605}       \\
BAE, MC-Dropout                            & N($\hat{x}^*$,1)                                               & 0.260            & 0.051       & 0.088       & 0.051       \\ \midrule
BAE, BayesBB                                    & Ber($\hat{x}^*$)                                             & 0.377            & 0.572       & 0.211       & 0.570       \\
BAE, BayesBB                                    & C-Ber($\hat{x}^*$)                                           & 0.385            & 0.591       & 0.193       & 0.586       \\
BAE, BayesBB                                    & N($\hat{x}^*$,1)                                               & 0.359            & 0.094       & 0.131       & 0.094       \\ \midrule
BAE, Ensemble                              & Ber($\hat{x}^*$)                                             & 0.186            & \textbf{0.603}       & 0.086       & \textbf{0.603}       \\
BAE, Ensemble                              & C-Ber($\hat{x}^*$)                                           & 0.175            & \textbf{0.609}       & 0.104       & \textbf{0.609}       \\
BAE, Ensemble                              & N($\hat{x}^*$,1)                                               & 0.151            & 0.039       & 0.065       & 0.039       \\ \bottomrule
\end{tabular}
}
\end{table}